\documentclass[conference]{IEEEtran}
\IEEEoverridecommandlockouts
\usepackage{cite}
\usepackage{amsmath,amssymb,amsfonts}
\usepackage{algorithmic}
\usepackage{graphicx}
\usepackage{textcomp}
\usepackage{xcolor}
\def\BibTeX{{\rm B\kern-.05em{\sc i\kern-.025em b}\kern-.08em
    T\kern-.1667em\lower.7ex\hbox{E}\kern-.125emX}}

\makeatletter
\newcommand{\newlineauthors}{%
  \end{@IEEEauthorhalign}\hfill\mbox{}\par
  \mbox{}\hfill\begin{@IEEEauthorhalign}
}
\makeatother

\begin{document}

\title{VAIS ASR: Building a conversational speech  recognition system using language model combination \\
\thanks{}
}

\author{\IEEEauthorblockN{1\textsuperscript{st} Quang Minh Nguyen}
\IEEEauthorblockA{\textit{Vietnam Artificial Intelligence System} \\
Hanoi, Vietnam \\
minhnq@vais.vn}
\and
\IEEEauthorblockN{2\textsuperscript{nd} Thai Binh Nguyen}
\IEEEauthorblockA{\textit{Vietnam Artificial Intelligence System} \\
\textit{Hanoi University of Science and Technology}\\
Hanoi, Vietnam \\
binhnguyen@vais.vn}
\newlineauthors
\IEEEauthorblockN{3\textsuperscript{rd} Ngoc Phuong Pham}
\IEEEauthorblockA{\textit{Vietnam Artificial Intelligence System} \\
\textit{Thai Nguyen University} \\
Thai Nguyen, Vietnam \\
phuongpn@tnu.edu.vn}
\and
\IEEEauthorblockN{4\textsuperscript{th} The Loc Nguyen}
\IEEEauthorblockA{
\textit{Vietnam Artificial Intelligence System} \\
\textit{Hanoi University of Mining and Geology} \\
Hanoi, Vietnam \\
locnguyen@vais.vn}
}

\maketitle

\begin{abstract}
Automatic Speech Recognition (ASR) systems have been evolving quickly and reaching human parity
in certain cases. The systems usually perform pretty well on reading style and clean speech, 
however, most of the available systems suffer from situation where the speaking style is
conversation and in noisy environments.
It is not straight-forward to tackle such problems due to difficulties in data collection for both
speech and text.
In this paper, we attempt to mitigate the problems using language models combination
techniques that allows us to utilize both large amount of writing style text and
small number of conversation text data. Evaluation on the VLSP 2019 ASR challenges showed that
our system achieved 4.85\% WER on the VLSP 2018 and 15.09\% WER on the VLSP 2019 data sets.
\end{abstract}


\section{Introduction}
Informal speech is different from formal speech,
especially in Vietnamese due to many conjunctive words in this language.
Building an ASR model to handle such kind of speech is particularly difficult
due to the lack of training data and also cost for data collection.
There are two components of an ASR system that contribute the most to the
accuracy of it, an acoustic model and a language model. While collecting data
for acoustic model is time-consuming and costly, language model data is
much easier to collect.

The language model training for the Automatic Speech Recognition (ASR) system usually based on corpus crawled on formal text, so that some conjunctive words
which often used in conversation will be missed out, leading to the
system is getting biased to writing-style speech.

In this paper, we present our attempt to mitigate the problems using a large
scale data set and a language model combination technique that only require
a small amount of conversation data but can still handle very well conversation
speech.

\section{System Description}
In this section, we describe our ASR system, which consists of 2 main components, 
an acoustic model which models the correlation between phonemes and speech signal;
and a language model which guides the search algorithm throughout inference process.

\subsection{Acoustic Model}
We adopt a DNN-based acoustic model \cite{peddinti2015time} with 11 hidden layers and 
the alignment used to train the model is derived from a HMM-GMM model trained with SAT criterion.
In a conventional Gaussian Mixture Model - Hidden Markov Model (GMM-HMM) acoustic model, the state emission log-likelihood of the observation feature vector $o_t$ for certain tied state $s_j$ of HMMs at time $t$ is computed as
\begin{equation}
\textmd{log } p(\mathbf{o}_t|s_j) = \textmd{log} \sum_{m=1}^M \pi_{jm}\mathbb{N}(\mathbf{o}_t|s_j )
\end{equation}
where M is the number of Gaussian mixtures in the GMM for state $j$ and $\pi_{jm}$ is the mixing weight. As the outputs from DNNs represent the state posteriors $p(s_j|\mathbf{o}_t)$, a DNN-HMM hybrid system uses pseudo log-likelihood as the state emissions that is computed as
\begin{equation}
\textmd{log } p(\mathbf{o}_t|s_j) = \textmd{log }  p(s_j|\mathbf{o}_t) - \textmd{log } p(s_j),
\end{equation}
where the state priors log $p(s_j )$ can be estimated using the state alignments on the training speech data.

\subsection{Language Model}
\begin{figure*}[htpb]
  \centering
  \includegraphics[width=0.8\linewidth]{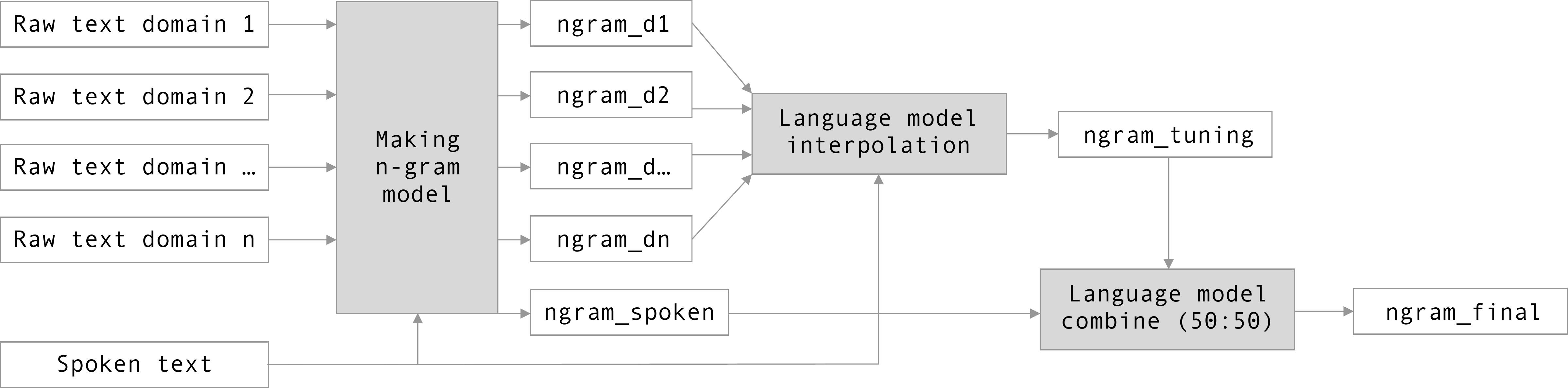}
  \caption{Language model training pipeline}
  \label{fig:lm_pipeline}
\end{figure*}

Our language model training pipeline is described in Figure~\ref{fig:lm_pipeline}.
First, we collect and clean large amount of text data
from various sources including news, manual labeled conversation video.
Then, the collected data is categorized into domains. 
This is an important step
as the ASR performance is highly depends on the speech domain.
After that, the text is fed into a data cleaning pipeline to clean bad
tone marks, normalizing numbers and dates.

For each domain text data, we train an n-gram language model\cite{brown1992class} that is
optimized for that domain. As the results, we have more than 10 language models. These language models are combined based on perplexity calculated
on a small text of a domain that we want to optimize for.

In our system, the language model is used in 2 pass-decoding.
In the first pass, the language model is combined with acoustic
and lexicon model to form a full decoding graph. In this stage,
the language model is typically small in size by utilizing pruning
method.
In the second stage, we use a un-pruned language model to
rescore decoded lattices.

\section{Corpus Description}
\subsection{Speech data}
Our speech corpus consists of approximately 3000 hours of speech data
including various domains and speaking styles. The data is augmented
with noise and room impulse respond to increase the quantity and
prevent over-fitting.

\subsection{Text data}
To train n-gram language models that are robust to various domains
and, we collect corpus from many resources, mainly is come from newspaper site (like dantri, vnexpress,..), law document and some crawled repository\footnote{The text corpus is made available here https://github.com/binhvq/news-corpus}. 
In total, more than 50GB of text split to separate subjects was used to train n-gram language models. Table \ref{lm_statistic} shows the statistic of the collected data.

\begin{table}[ht]
\small
\centering
\caption{Language model dataset}
\begin{tabular}{|c|c|}
    \hline  
    \textbf{Domain} & \textbf{Vocab size} \\
    \hline
    \hline
    Cong nghe & 269k \\
    \hline
    Doi song & 285k \\
    \hline
    Giai tri & 305k \\
    \hline
    Giao duc & 135k \\
    \hline
    Khoa hoc & 167k \\
    \hline
    Kinh te & 291k \\
    \hline
    Phap luat & 446k \\
    \hline
    Tin tuc & 1.247k \\
    \hline
    Nha dat & 24.97 \\
    \hline
    The gioi & 126k \\
    \hline
    The thao & 216k \\
    \hline
    Van hoa & 300k \\
    \hline
    Xa hoi & 203k \\
    \hline
    Xe co & 92k \\
    \hline
\end{tabular}
\label{lm_statistic}
\end{table}






\section{Experiments}
There are two different testing sets from VLSP 2018 and VLSP 2019. In general, the data of this year is more complex than the last year one, so there is a big gap in results between two of them. The experiments are conducted using the Kaldi speech recognition toolkit\cite{povey2011kaldi}.

\subsection{Evaluation data}
\begin{itemize}
    \item \textit{Testing data VLSP 2018: }This dataset contains relatively 5 hours of short audio voices with 796 samples. 
    \item \textit{Testing data VLSP 2019: }The testing set in 2019 is quite harder compare with the set in 2018. There are more than 17 hours with 16,214 short audio samples. The set is difficult because audio contains more noise and having more informal speaking style. 
\end{itemize}

\subsection{Single system evaluation}
Table~\ref{tab:result_vlsp2019_news} shows the results on the
VLSP 2019 test set with two language models.
Various language model weights were also used to find the optimal value
for the test set. As we can see, the conversation language model
with the weight of 8 yielded the best single system of 15.47\% WER.

\begin{table}[ht]
\small
\centering
\caption{Evaluation of the system with different language models}
\begin{tabular}{|c|c|c|}
    \hline  
    LMWT & General LM & Conversation LM \\
    \hline
    7 & 18.85 & 15.67 \\
    \hline
    8 & 20.05 & \textbf{15.47} \\
    \hline
    9 & 22.11 &  15.78 \\
    \hline
    10 & 24.97 & 16.72 \\
    \hline
\end{tabular}
\label{tab:result_vlsp2019_news}
\end{table}





\subsection{System combination}
To further improve the performance, we adopt system combination
on the decoding lattice level. By combining systems, we can take advantage
of the strength of each model that is optimized for different
domains.
The results for 2 test sets is showed on Table~\ref{tab:vlsp2018_combine} and \ref{tab:vlsp2019_combine}.

As we can see, for both test sets, system combination significantly
reduce the WER. The best result for vlsp2018 of 4.85\% WER is obtained by the
combination weights 0.6:0.4 where 0.6 is given to the general language model and 0.4 is given to the conversation one.
On the vlsp2019 set, the ratio is change slightly by 0.7:0.3 to
deliver the best result of 15.09\%.

\begin{table}[ht]
\small
\centering
\caption{Evaluation in Word Error Rate (WER) with VLSP 2018 set}
\begin{tabular}{|c|c|c|c|}
    \hline  
    LMWT & General LM ratio & Conversation LM ratio & WER \\
    \hline
    7 & 0.3 & 0.7 & 5.08\% \\
    \hline
    7 & 0.4 & 0.6 & 5.02\% \\
    \hline
    7 & 0.5 & 0.5 & 4.94\% \\
    \hline
    7 & 0.6 & 0.4 & 4.86\% \\
    \hline
    7 & 0.7 & 0.3 & 4.90\% \\
    \hline
    8 & 0.3 & 0.7 & 5.12\% \\
    \hline
    8 & 0.4 & 0.6 & 5.02\% \\
    \hline
    8 & 0.5 & 0.5 & 4.90\% \\
    \hline
    \textbf{8} & \textbf{0.6} & \textbf{0.4} & \textbf{4.85\%} \\
    \hline
    8 & 0.7 & 0.3 & 4.93\% \\
    \hline
    9 & 0.3 & 0.7 & 5.26\% \\
    \hline
    9 & 0.4 & 0.6 & 5.17\% \\
    \hline
    9 & 0.5 & 0.5 & 5.04\% \\
    \hline
    9 & 0.6 & 0.4 & 5.07\% \\
    \hline
    9 & 0.7 & 0.3 & 5.09\% \\
    \hline
    10 & 0.3 & 0.7 & 5.64\% \\
    \hline
    10 & 0.4 & 0.6 & 5.52\% \\
    \hline
    10 & 0.5 & 0.5 & 5.37\% \\
    \hline
    10 & 0.6 & 0.4 & 5.37\% \\
    \hline
    10 & 0.7 & 0.3 & 5.37\% \\
    \hline
\end{tabular}
\label{tab:vlsp2018_combine}
\end{table}
    \begin{table}[ht]
\small
\centering
\caption{Evaluation in WER with VLSP 2019 set}
\begin{tabular}{|c|c|c|c|}
    \hline  
    LMWT & General LM ratio & Conversation LM ratio & WER \\
    \hline
    7 & 0.5 & 0.5 & 15.55\% \\
    \hline
    7 & 0.6 & 0.4 & 15.18\% \\
    \hline
    7 & 0.7 & 0.3 & 15.15\% \\
    \hline
    7 & 0.8 & 0.2 & 15.26\% \\
    \hline
    8 & 0.5 & 0.5 & 15.88\% \\
    \hline
    8 & 0.6 & 0.4 & 15.27\% \\
    \hline
    \textbf{8} & \textbf{0.7} & \textbf{0.3} & \textbf{15.09\%} \\
    \hline
    8 & 0.8 & 0.2 & 15.10\% \\
    \hline
    9 & 0.5 & 0.5 & 16.83\% \\
    \hline
    9 & 0.6 & 0.4 & 16.06\% \\
    \hline
    9 & 0.7 & 0.3 & 15.67\% \\
    \hline
    9 & 0.8 & 0.2 & 15.55\% \\
    \hline
    10 & 0.5 & 0.5 & 18.47\% \\
    \hline
    10 & 0.6 & 0.4 & 17.40\% \\
    \hline
    10 & 0.7 & 0.3 & 16.85\% \\
    \hline
    10 & 0.8 & 0.2 & 16.60\% \\
    \hline
\end{tabular}
\label{tab:vlsp2019_combine}

\end{table}

\section{Conclusion}
In this paper, we presented our ASR system participated in VLSP 2019 challenge that incorporates a language model combination technique to handle conversation speech with
small amount of text data required.
The method demonstrated that it can help to reduce WER by 3\% on the VLSP 2019 challenge.
\label{Bibliography}
\bibliographystyle{IEEEtran} 
\bibliography{ref} 

\end{document}